\documentclass[lettersize,journal]{IEEEtran}
\usepackage{amsmath,amsfonts}
\usepackage{algorithmic}
\usepackage{array}
\usepackage[caption=false,font=normalsize,labelfont=sf,textfont=sf]{subfig}
\usepackage{textcomp}
\usepackage{stfloats}
\usepackage{url}
\usepackage{verbatim}
\usepackage{tabularx}
\usepackage{graphicx}
\usepackage{adjustbox}
\usepackage{booktabs}
\usepackage{etoolbox}
\makeatletter
\patchcmd{\@makecaption}
  {\scshape}
  {}
  {}
  {}
\makeatletter
\patchcmd{\@makecaption}
  {\\}
  {.\ }
  {}
  {}
\makeatother

\usepackage{colortbl,xcolor}
\definecolor{amethyst}{rgb}{0.6, 0.4, 0.8}
\usepackage{multirow}
\usepackage{multicol}

\newcommand{\cmark}{\ding{51}}%
\newcommand{\xmark}{\ding{55}}%
\usepackage{makecell}
\usepackage{epsfig}
\usepackage{amssymb}
\usepackage{mathtools}
\usepackage{pifont}

\usepackage{enumitem}
\usepackage{multirow}
\usepackage{wrapfig}
\usepackage{colortbl,xcolor}
\usepackage{mathtools}

\setlist[itemize]{align=parleft,left=0pt}

\definecolor{azure(colorwheel)}{rgb}{0.0, 0.5, 1.0}
\definecolor{nicegreen}{rgb}{0.0, 0.7, 0.1}
\definecolor{customgray}{rgb}{0.9, 0.9, 0.9}
\definecolor{pink}{cmyk}{0, 0.7808, 0.4429, 0.1412}
\definecolor{amethyst}{rgb}{0.6, 0.4, 0.8}
\definecolor{black}{rgb}{0.0, 0.0, 0.0}
\definecolor{white}{rgb}{1.0, 1.0, 1.0}
\definecolor{blue}{rgb}{0.0, 0.0, 0.4}
\definecolor{red}{rgb}{0.4, 0.0, 0.0}

\usepackage[font={small,color=black}]{caption}

\newcolumntype{g}{>{\columncolor{customgray}}c}
\newcolumntype{z}{>{\columncolor{customgray}}l}
\newcolumntype{?}[1]{!{\vrule width #1}}

\renewcommand{\paragraph}[1]{\vspace{0.4mm}\noindent\textbf{#1.}\,\,}
\usepackage{xspace}

\def\onedot{.\@\xspace}
\def\ie{\emph{i.e}\onedot}

\def\etc{\emph{etc}\onedot}

\def\etal{\emph{et al}\onedot}

\newcommand{\be}{\begin{eqnarray}}
\newcommand{\ee}{\end{eqnarray}}
\newcommand{\bee}{\begin{eqnarray*}}
\newcommand{\eee}{\end{eqnarray*}}

\newcommand{\matrixb}{\left[ \begin{array}}
\newcommand{\matrixe}{\end{array} \right]}

\usepackage[
]{caption}
\def\BibTeX{{\rm B\kern-.05em{\sc i\kern-.025em b}\kern-.08em
    T\kern-.1667em\lower.7ex\hbox{E}\kern-.125emX}}

\usepackage{balance}
\begin{document}
\title{The Devil in the Details: Simple and Effective Optical Flow Synthetic Data Generation}
\author{Kwon Byung-Ki$^{2}$, Kim Sung-Bin$^{1}$, Tae-Hyun Oh$^{1,2}$\\
$^{1}$Department of Electrical Engineering, POSTECH $\quad$ $^{2}$Graduate School of AI, POSTECH \\ 
\texttt{\{byungki.kwon, sungbin, taehyun\}@postech.ac.kr}
}

\maketitle

\begin{abstract}
Recent work on dense optical flow has shown significant progress, primarily in a supervised learning manner requiring a large amount of labeled data. Due to the expensiveness of obtaining large scale real-world data, computer graphics are typically leveraged for constructing datasets. However, there is a common belief that synthetic-to-real domain gaps limit generalization to real scenes. In this paper, we show that the required characteristics in an optical flow dataset are rather simple and present a simpler synthetic data generation method that achieves a certain level of realism
with compositions of elementary operations. With 2D motion-based datasets, we systematically analyze the simplest yet critical factors for generating synthetic datasets. Furthermore, we propose a novel method of utilizing occlusion masks in a supervised method and observe that suppressing gradients on occluded regions serves as a powerful initial state in the curriculum learning sense.
The RAFT network initially trained on our dataset outperforms the original RAFT on the two most challenging online benchmarks, MPI Sintel and KITTI 2015.
\end{abstract}

\section{Introduction}\label{sec:introduction}
Optical flow provides the clues of motion between subsequent frames, which can be utilized for other computer vision tasks such as object tracking, action recognition, 3D reconstruction, and video enhancement, \etc Recently, deep neural networks have shown great progress in optical flow estimation~\cite{flownet2,pwc,sun2019models,liteflownet,raft}. 
The progress has been made primarily in a supervised learning manner requiring a large amount of labeled data.
Despite the effectiveness of the learning-based approaches, obtaining labeled real-world data is prohibitively expensive at a large scale. 
Therefore, synthetic computer graphics data~\cite{flownet,dcoco,things,sintel} are typically leveraged.

A common belief of using synthetic data is that the data rendered by graphics engines limit generalization to real scenes due to synthetic-to-real domain gaps in quality.
Those gaps involve real-world effects such as noise, 3D motion, non-rigidity, motion blur, occlusions, large displacements, and texture diversity. 
Thus, synthetic datasets~\cite{flownet,dcoco,things,sintel} for optical flow have been developed by considering these effects to some extent, \ie, mimicking the real-world effects.

In this paradigm, we throw a question,
``Which factor of the synthetic dataset is essential for the generalization ability to the real domain?'' In this work, we found that the required characteristics for an optical flow dataset are simple; achieving only a certain level of realism is enough for training highly generalizable and accurate optical flow models.
We empirically observe that a simple 2D motion-based dataset as training data often shows favorable performance for ordinary purposes or much higher than the former
synthetic datasets~\cite{dcoco,mayer2018makes}, which are rendered by complex 3D object or motion with rich textures. 
Furthermore, we found that using occlusion masks to give the network incomplete information is effective for a powerful initial state of curriculum learning.
\begin{figure*}[t]
\centering
\resizebox{0.95\linewidth}{!}{
\begin{tabular}{@{}c@{}c@{\,}|@{\,}c@{}c@{\,}|@{\,}c@{}c@{\,}|@{\,}c@{}c@{}}
\includegraphics[width=0.24\textwidth]{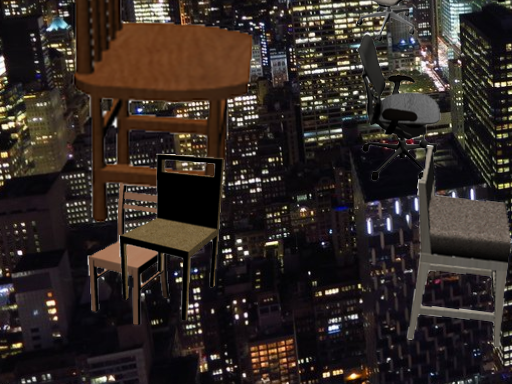}&
\includegraphics[width=0.24\textwidth]{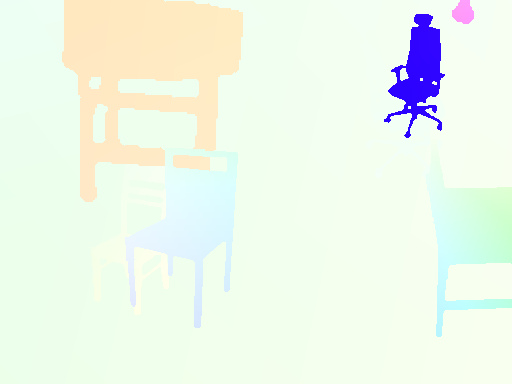}&
\includegraphics[width=0.24\textwidth]{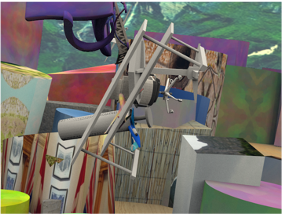}&
\includegraphics[width=0.24\textwidth]{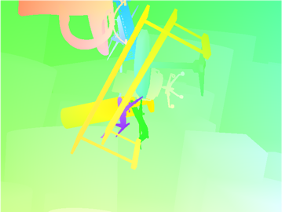}&
\includegraphics[width=0.24\textwidth]{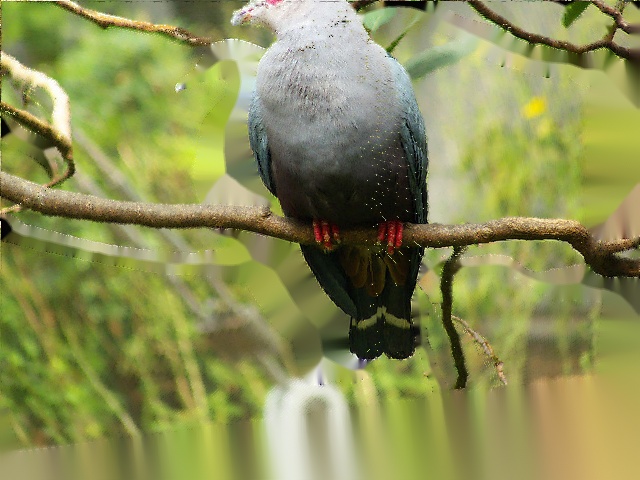}&
\includegraphics[width=0.24\textwidth]{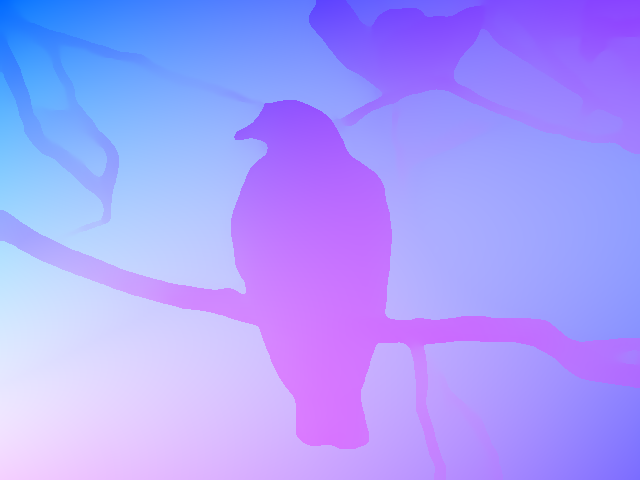}&
\includegraphics[width=0.24\textwidth]{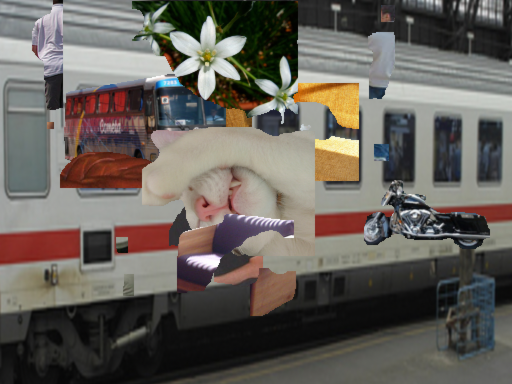}&
\includegraphics[width=0.24\textwidth]{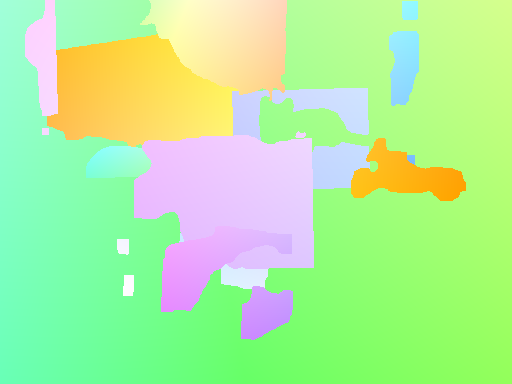}\\[1mm]
\multicolumn{2}{@{}c@{\,}|@{\,}}{\Large (a) FlyingThings3D~\cite{things}}&
\multicolumn{2}{@{}c@{\,}|@{\,}}{\Large (b) FlyingChairs~\cite{flownet}}&
\multicolumn{2}{@{}c@{\,}|@{\,}}{\Large (c) dCOCO~\cite{dcoco}}&
\multicolumn{2}{@{}c@{}}{\Large (d) Ours}
\end{tabular}
}
    \caption{
    \textbf{The prior arts of synthetic data and our proposed dataset.} 
    Sampled frames and its corresponding flow maps are visualized.
    While being diverse in motion, (a,b) include many thin object parts and unrealistically simple reflectance.
    (c) includes semantically coherent flow map but the diversity of the motion is limited by a global camera motion. 
    Our method, in contrast, includes both controllable and diverse motion characteristics with semantically coherent object shapes and rich texture.
    }
    \vspace{-3mm}
    \label{figure:1}
\end{figure*}

We design easily controllable synthetic dataset generation recipes using a cut-and-paste method with segmented 2D object textures.
As shown in Fig.~\ref{figure:1}, our generated data appears to be far from the real-world one, but training on those shows promising results both on generalization and fine-tuning regimes, outperforming the networks trained on the competing datasets. We also utilize occlusion masks to stop gradients on occluded regions, and the RAFT network initially trained with occlusion masks outperforms the original RAFT on the two most challenging online benchmarks, MPI Sintel~\cite{sintel} and KITTI 2015~\cite{kitti15}.
Our key contributions are summarized as follows:
\begin{itemize}
    \item We present simple synthetic data generation recipes with compositions of simple elementary operations and show comparable performance against competing methods.
    \item We propose a novel method of utilizing occlusion masks in a supervised method and show that suppressing gradients on occluded regions in a supervised optical flow serves as a powerful initial state in the curriculum learning protocol.
    \item We systematically analyze our dataset and the effects according to different factors of motion type, motion distribution, data size, texture diversity, and occlusion masks.
\end{itemize}

\section{Related Work}\label{sec:relatedwork}

We briefly review our target task, \ie, optical flow estimation, and the training datasets that have been used for training learning-based optical flow estimation methods.

\paragraph{Optical Flow} 
Fundamentally, optical flow estimation for each pixel is an ill-posed problem. Traditional approaches~\cite{horn_and_shcunk, Black_M.J., TV-1, discrete} attempted to deal with imposing smoothness priors to regularize the ill-condition in an optimization framework. 
According to the advance of deep learning, the ill-posedness has been tackled by learning, yielding superior performance.
Starting with the success of FlowNet~\cite{flownet,flownet2}, recent
optical flow estimation methods have been developed by supervised learning~\cite{Volumetric,costvolume,pwc,raft,Iterative_residual}.
However, these approaches strongly rely on training datasets, where 
real supervised data of optical flow is extremely difficult to obtain~\cite{mayer2018makes}.

\begin{figure}[t]
\centering
\resizebox{1.0\linewidth}{!}{
\begin{tabular}{@{}c|c@{}}
\includegraphics[width=0.565\textwidth]{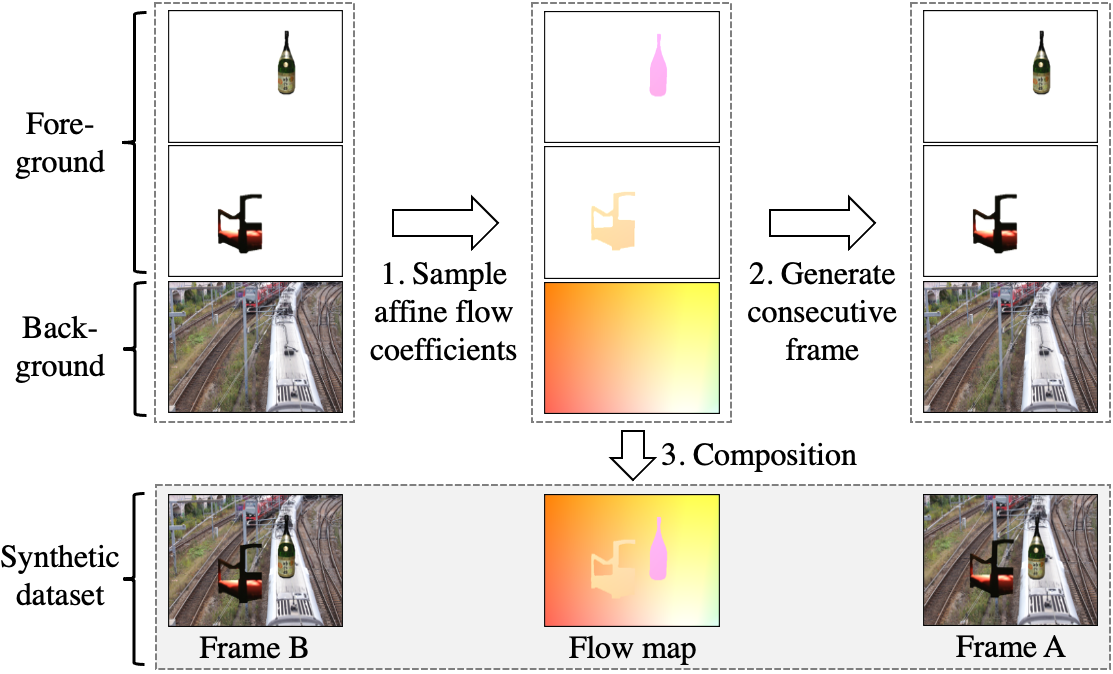}&
\includegraphics[width=0.45\textwidth]{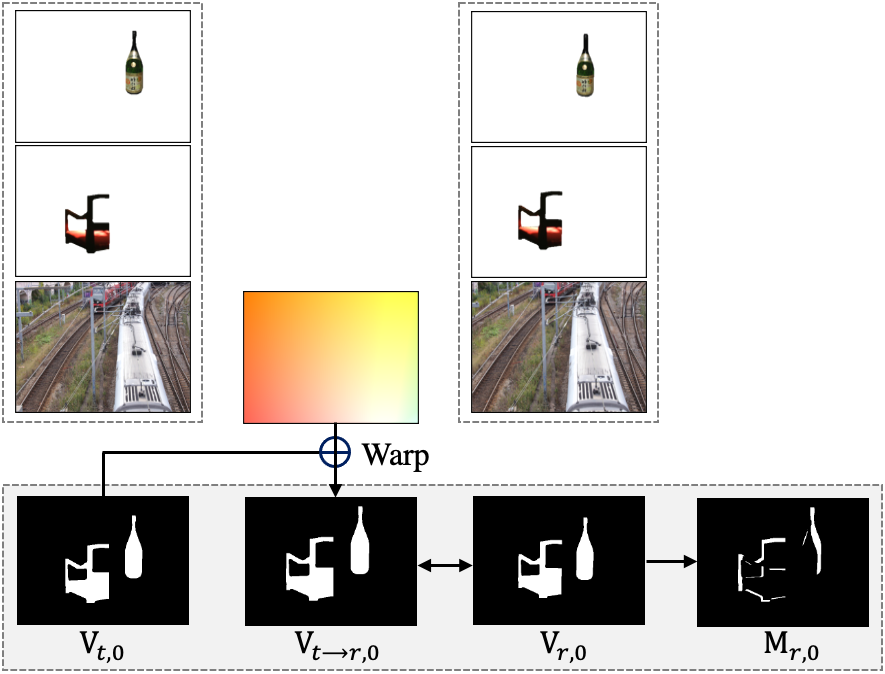}\\[0mm]
\small (a) Data generation pipeline & \small (b) Occlusion mask estimation
\end{tabular}
}
    \caption{
    \textbf{Schematic overview of our data generation pipeline and occlusion mask estimation.} (a) Given a background image and foreground objects, we sample affine flow coefficients and generate a consecutive frame. These coefficients can be used to extract exact ground-truth optical flow map. (b) We describe the process of estimating the occlusion mask ($\mathrm{M}_{r,i}$) for the first layer ($i=0$), which is the background. This process is recursively conducted in ascending order until the end of the layers.}
    \vspace{-3mm}
    \label{figure:2}
\end{figure}

\paragraph{Datasets} 
The supervised learning-based methods for optical flow estimation requires exact and pixel-accurate ground truth.
While obtaining true real motion is extremely difficult without the support of additional information, several real-world optical flow datasets~\cite{kitti12,kitti15,slowflow,HD1k} have been proposed.
However, these datasets are relatively small scale and biased to limited scenarios; thus, those are not sufficient for training a deep model but more suitable for benchmark test sets.

To address persistent data scarcity, studies for generating large-scale synthetic datasets have been attempted. 
Dosovitskiy~\etal~\cite{flownet} propose a synthetic dataset of moving 3D chairs superimposed on the images from Flickr. 
Similarly, Mayer~\etal~\cite{things} present datasets where not only chairs but various objects are scattered in the background.
Aleotti~\etal~\cite{dcoco} leverage an off-the-shelf monocular depth network to synthesize a novel view from a single image and compute an accurate flow map. 

Mayer~\etal~\cite{mayer2018makes} present critical factors of the synthetic dataset, \ie, the object shape, motion types and distributions, textures, real-world effects, data augmentation, and learning schedules. Sun~\etal~\cite{autoflow} generate a learning-based synthetic dataset for training accurate optical flow networks, but it is still challenging to distinguish the key factors for synthetic data intuitionally.
We build upon the observations of Mayer~\etal~\cite{mayer2018makes} and design easily controllable synthetic dataset generation recipes and identify additional key factors such as \emph{balanced motion distribution, amount of data, texture combination, and learning schedules with occlusion masks}.

\section{Data Generation Pipeline}\label{sec:pipeline}
In this section, we present a simple method to generate an effective optical flow dataset.
Unlike the prior arts using 3D motions and objects with computer graphics, 
our generation scheme remains simple by using 2D image segment datasets and 2D affine motion group. 
The proposed simple dataset enables analyzing the effect of each factor of the synthetic dataset.

\paragraph{Overall Pipeline} 
The overall data generation pipeline is illustrated in Fig.~\ref{figure:2}.
As shown, we use a simple cut-and-paste method where foreground objects are pasted on an arbitrary background image. 
Inspired by Oh~\etal~\cite{mm}, the segmented foreground objects and random background images are obtained from two independent datasets to encourage combinatorial diversity while avoiding texture overlaps. In this work, we use PASCAL VOC~\cite{voc} and MS COCO~\cite{mscoco} as suggested by Oh~\etal~\cite{mm}.
The foreground objects are first superimposed randomly, and its consecutive frame is composed of randomly moving both the foreground objects and the background image by simple affine motions.
This allows us to express diverse motions, easily control the motion distribution, and compute occlusion masks. 

\paragraph{Background Processing} 
We first sample an image from an image dataset for background and resize them to $712\times 584$.
We regard this frame as the target frame (Frame B in Fig.~\ref{figure:2}). Then, we generate a flow map using random affine coefficients, including translation, rotation, and scaling (zooming), and inverse-warp the target frame to obtain the reference frame (Frame A in Fig.~\ref{figure:2}).
We sample the translation coefficient of background from the range $[-20, 20]$ pixels for each direction, and with a $30$\% chance, the translation coefficient is reset to zero.
The rotation and scale coefficients are sampled from $[-\frac{\pi}{100}$,$ \frac{\pi}{100}]$ and $[0.85, 1.15]$, respectively. 
From the sampled affine matrix, we obtain a ground-truth flow map by subtracting the coordinates of two background image pairs as $\mathbf{f} = \mathbf{A}\mathbf{x}-\mathbf{x}$, where $\mathbf{f}$ denotes each flow vector of a pixel at the reference frame, $\mathbf{A}$ the affine transform, and $\mathbf{x}$ a homogeneous coordinate $[x,y,1]$ of each pixel on the reference frame.
We sample 7,849 background images from MS COCO~\cite{mscoco}.

\paragraph{Foreground Processing}
For synthesizing foreground objects' motion, we use segmented objects from a semantic image segmentation dataset.
For the target frame, we first sample the number of foreground objects to be composited in $\{7,8,\cdots,14,15\}$. 
Then, we randomly place these objects on the target one and apply inverse-warping to obtain the warped objects on the reference frame using optical flow maps obtained from random affine transformations.
The sampling ranges of rotation and scale coefficients are the same as those of the background case.
The distribution of the translation coefficient is designed to follow the exponential distribution as $\tfrac{1}{Z}\exp(-f/T)$, where the temperature $T$ is empirically set to $20$, and $Z$ the normalization term. The distribution is inspired by natural statistics of optical flow~\cite{roth2007spatial}, where the statistics of motions tend to follow Laplacian distribution.
We limit the distribution range $[0,150]$ by resampling if the magnitude is over $150$ pixels. 
The translation direction of foregrounds is sampled at uniformly random. We use $2913$ images from PASCAL VOC~\cite{voc}, and from the set, we extract $5543$ preprocessed segments as foreground objects.

\paragraph{Composition} 
We sequentially paste foregrounds on the background to generate a single pair of consecutive frames. 
We particularly care about the regions near object boundaries when compositing optical flow maps.
Directly applying alpha blending of a stack of foreground flow maps with a background flow map yields inconsistent flows near object boundaries.
To deal with this, we paste the flow maps of each foreground only when the alpha channel value is at least $0.4$. 
After composition, we conduct the center crop to the composited images to obtain outputs of size $512\times384$\, which is the same as FlyingChairs~\cite{flownet}. 
Our data generation speed is faster than AutoFlow~\cite{autoflow}, which generates a learning-based dataset for given target data, and ours
about 500 times faster than dCOCO~\cite{dcoco} as shown in Table~\ref{table1}. Our fast data generation speed is beneficial for analyzing the required characteristics to train accurate optical flow networks.

\paragraph{Occlusion Mask}
Similar to the prior arts~\cite{things,sintel,kitti12,kitti15}, our data generation method exports occlusion masks as well.
Predicting motions of regions being occluded is an intractable problem and requires uncertain forecasting, which can act as detrimental outliers during training. Thus, prior arts~\cite{hur2019iterative,jeong2022imposing} estimate occlusion masks as well to encourage reliable optical flow estimation.
Unlike prior arts, we utilize occlusion masks in a supervised method by suppressing gradients on
occluded regions in a supervised optical flow. The gradient suppression with occlusion masks serves as a powerful initial state in the curriculum learning protocol, which will be discussed in the experimental section.
To obtain occlusion masks, given the alpha maps of each layer
including foregrounds ($i\geq 1$) and background ($i=0$) in order, we binarize the alpha map by thresholding with $0.4$, denoting $\mathrm{\alpha}_{\{r, t\},i}$ for the $i$-th object layer in the reference and target frames, respectively.
The non-visible regions $\mathrm{V}_{\{r, t\},i}$ of the $i$-th layer in each frame are computed by
$\mathrm{V}_{\{r, t\},i} = \alpha_{\{r, t\},i}{\cap}(\cup_{k=i+1}^{L}{\alpha_{\{r, t\},k}})$.
Using the $i$-th layer flow map $\mathbf{f}_{i}$,
we inverse-warp the $\mathrm{V}_{t,i}$ to the reference frame as $\mathrm{V}_{t\rightarrow r,i} = \mathbf{f}_{i}\circ\mathrm{V}_{t,i}$ and binarize it by $0.4$ again, where $\circ$ denotes the warping operation.
Then, because the occluded regions are only visible in the 
reference frame, we can find such an occlusion mask of each layer 
by $\mathrm{M}_{r,i} = \mathrm{max}(\mathrm{V}_{t\rightarrow r,i} - \mathrm{V}_{r,i}, 0)$.
The compromised occlusion mask $\mathrm{M}_{r}$ is obtained by
$\mathrm{M}_{r} = \cup_{i=0}^{L} \mathrm{M}_{r,i}$.

\section{Experiments}
In this section, we compare the performance of respective optical flow networks by training on our datasets with/without the occlusion mask and competing datasets. Utilizing the simple data generation recipe, we also 
analyze the effects of characteristics in optical flow datasets.

\begin{table}
\caption{\textbf{Data generation speed.} We evaluate the speed for generating 100 pairs of synthetic data with a single NVIDIA Titan RTX GPU: (A) dCOCO, and (B, C, D) ours with the different number of foregrounds. The number of the foregrounds is sampled between 7 to 15.}
    \centering
    \label{table1}
    \resizebox{\linewidth}{!}{
    \begin{tabular}{cccc}
        \toprule
        & Dataset & Number of& 100 pairs \\
        &  & foregrounds & generation time \\
        \midrule 
        (A) & AutoFlow~\cite{autoflow} & - & 336 days \\
        (B) & dCOCO~\cite{dcoco} & - & 5593.2 s \\
        (C) & Ours & 2 & 6.86 s\\
        (D) & Ours & 7 & 9.49 s\\
        (E) & Ours & 15 & 12.98 s\\
        \bottomrule
    \end{tabular}
    }
\end{table}
\begin{table*}[t]
\footnotesize
\centering
    \caption{\textbf{Comparison with other datasets.} We evaluate the generalization and fine-tuning test of the RAFT networks trained on training datasets: (A) FlyingChairs, (B) dCOCO, (C) ours, (D) ours with occlusion mask, (E) FlyingChairs and FlyingThings3D, (F) ours and FlyingThings3D, and (G) ours with occlusion mask and FlyingThings3D. (B$\dagger$) is obtained from the original paper of \cite{dcoco}.}
    \label{table2}
    \resizebox{\linewidth}{!}{
    \begin{tabular}{@{}c@{}@{}c@{}@{}c@{}}
    \begin{tabular}{c@{\,\,\,\,\,\,}c@{\,\,\,\,\,}c@{\hspace{0.5cm}}}
    \toprule
    &\multicolumn{2}{c}{}\\[5pt]
    &\multicolumn{2}{c}{}\\[5pt]
    & Dataset & Motions\\
    \hline
    (A) & Ch & 2D\\
    (B$\dagger$) & dCOCO & 3D\\
    (B) & dCOCO & 3D\\
    \hline
    (C) & Ours & 2D  \\
    (D) & Ours+O & 2D    \\
    \midrule
    (E) &  Ch $\rightarrow$ Th & 2D+3D\\
    \hline
    (F) & Ours$\rightarrow$Th & 2D+3D  \\
    (G) & Ours+O$\rightarrow$Th & 2D+3D  \\
    \bottomrule
    \end{tabular}
    &
    \begin{tabular}{c@{\quad}cc@{\quad}cc@{\quad}cc@{\quad}c@{\hspace{0.5cm}}}
    \toprule
    \multicolumn{8}{c}{Generalization test}\\
    \cmidrule(l{2pt}r{13pt}){1-8}
    \multicolumn{2}{c}{Sintel C.} & \multicolumn{2}{c}{Sintel F.} & \multicolumn{2}{c}{KITTI12} & \multicolumn{2}{c@{\hspace{0.5cm}}}{KITTI15} \\ 
    \cmidrule(lr){1-2}\cmidrule(lr){3-4}\cmidrule(lr){5-6}\cmidrule(l{2pt}r{13pt}){7-8}
    EPE & $\leq$1 & EPE & $\leq$1 & EPE & Fl & EPE & Fl \\
    \hline
    2.28 & 0.79 & 4.51 & 0.72 & 4.66 & 30.54 & 9.85 & 37.56 \\ 
    - & - & - & - & - & - & - & - \\
    2.62 & 0.45 &  3.90 & 0.39 & \textbf{1.82} & \textbf{6.62} & \textbf{3.81} &
    \textbf{12.43} \\
    \hline
    \textbf{1.98} & \textbf{0.86} & 3.85 & \textbf{0.82} & 3.63 & 20.00 & 7.17 & 29.24 \\
    2.02 & \textbf{0.86} & \textbf{3.67} & \textbf{0.82} & 3.66 & 19.37 & 7.88 & 28.41 \\
    \midrule
    1.47 & 0.90 & \textbf{2.79} & 0.85 & 2.15 & 9.30 & 5.00 & 17.44 \\
    \hline
    \textbf{1.29} & \textbf{0.91} & 2.81 & 0.85 & 2.04 & 9.02 & \textbf{4.77} & 16.72 \\
    \textbf{1.29} & \textbf{0.91} & 2.86 & \textbf{0.86} & \textbf{2.03} & \textbf{8.64} & 4.84 & \textbf{16.38} \\
    \bottomrule
    \end{tabular}
    &
    \begin{tabular}{c@{\quad}cc@{\quad}cc@{\quad}cc@{\quad}c}
    \toprule
    \multicolumn{8}{c}{Finetuning test}\\
    \cmidrule(lr){1-8}
    \multicolumn{2}{c}{Sintel C.} &
    \multicolumn{2}{c}{Sintel F.} &
    \multicolumn{2}{c}{KITTI12} & \multicolumn{2}{c}{KITTI15} \\
    \cmidrule(lr){1-2}\cmidrule(lr){3-4}\cmidrule(lr){5-6}\cmidrule(lr){7-8}
    EPE & $\leq$1 & EPE & $\leq$1 & EPE & Fl & EPE & Fl \\
    \hline
    0.89 & 0.93 & 1.49 & 0.89 & 1.39 & 4.69 & 2.36 & 8.43 \\
    - & - & - & - & 1.37 & 4.70 & 2.76 & 9.15 \\
    1.08 & 0.92 & 1.84 & 0.88 & 1.37 & 4.76 & 2.57 & 8.81 \\
    \hline
    \textbf{0.85} & \textbf{0.94} & 1.40 & \textbf{0.89} & \textbf{1.33} & 4.37 & 2.20 & 8.19 \\
    0.89 & 0.93 & \textbf{1.39} & \textbf{0.89} & 1.35 & \textbf{4.36} & \textbf{2.15} & \textbf{7.60} \\
    \midrule
    0.84 & 0.93 & 1.31 & 0.89 & 1.31 & 4.25 & 2.28 & 7.96 \\
    \hline
    \textbf{0.83} & \textbf{0.94} & \textbf{1.29} & \textbf{0.90} & 1.32 & 4.24 & 2.10 & 7.52 \\
    0.86 & \textbf{0.94} & 1.31 & \textbf{0.90} & \textbf{1.28} & \textbf{4.11} & \textbf{2.02} & \textbf{7.34}\\
    \bottomrule
    \end{tabular}
    \end{tabular}}
\end{table*}

\begin{table}
\footnotesize
\centering
     \caption{\textbf{Generalization results on other benchmarks.} We evaluate the generalization test of the RAFT networks on HD1K~\cite{HD1k} and Virtual KITTI~\cite{vkitti}.}
    \label{table3}
    \resizebox{5.5cm}{!}{
    \begin{tabular}{cccc}
    \toprule
    & Dataset & HD1K & Virtual KITTI\\
    & & (real) & (synthetic) \\ 
    \midrule
    (A) & Ch &  1.70 & 6.52 \\
    (B) & dCOCO & 1.44 & \textbf{3.92} \\
    (C) & Ours & \textbf{1.06} & 6.38 \\
    \bottomrule
    \end{tabular}}
\end{table}

\paragraph{Optical Flow Network}
We use RAFT~\cite{raft} as a reference model to evaluate the benefits of our synthetic dataset in generalization and fine-tuning setups.
RAFT is a representative supervised model that is widely used to estimate the effectiveness of optical flow datasets~\cite{autoflow,dcoco}.
We follow the same hyper-parameters suggested by the implementation of 
Teed~\etal~\cite{raft}, and the experiment setup by 
Aleotti~\etal~\cite{dcoco} that shows one-/multi-stage training results.
For our synthetic datasets, in the initial training stage, we train RAFT for $100$k iterations with the batch size\footnote{The authors of \cite{raft,dcoco} use the batch size of $12$ and $6$ for training FlyingChairs and dCOCO, respectively.} of $10$, image crops of size $496\times 368$, the learning rate $4\times10^{-4}$, and the weight decay of $1\times10^{-4}$.

For multi-stage training with FlyingThings3D~\cite{things}, from the RAFT networks pre-trained on our datasets, we further train with the \texttt{frames\_cleanpass} split of FlyingThings3D that includes 40k consecutive frame pairs.
We train the model for $100$k iterations with a batch size of $6$, image crops of size $720\times 400$, the learning rate of $1.25\times10^{-4}$, and the weight decay of $1\times10^{-4}$.
These hyper-parameters are the same with the \emph{Things training stage} reported in 
\cite{raft}.

\begin{table}[t]
    \centering 
    \caption{\textbf{Test results on Sintel and KITTI 2015.} We evaluate the test performance of RAFT and RAFT-ours. Using our synthetic dataset with occlusion masks as an initial learning schedule achieves the higher performance in Sintel and KITTI 2015 test set. 
    }
    \label{table4}
    \resizebox{\linewidth}{!}{
    \begin{tabular}{ccccccc}
    \toprule
    & & \multicolumn{2}{c}{w/\textit{warm-start}} & \multicolumn{2}{c}{wo/\textit{warm-start}} & - \\
    \cmidrule(){3-4}\cmidrule(lr){5-6}\cmidrule(lr){7-7}
    & & Sintel C. & Sintel F. & Sintel C. & Sintel F. & KITTI15 \\ 
    \cmidrule(lr){3-3}\cmidrule(lr){4-4}\cmidrule(lr){5-5}\cmidrule(lr){6-6}\cmidrule(lr){7-7}
    & Training methods & EPE & EPE & EPE & EPE & Fl \\
    \midrule
    (A) &RAFT & 1.61 & 2.86 & 1.94 & 3.18 & 5.1 \\ 
    (B) &RAFT-Ours+O & \textbf{1.59} & \textbf{2.83} & \textbf{1.81} & \textbf{3.10} & \textbf{4.91}  \\
    \bottomrule
    \end{tabular}
    }
\end{table}

\begin{table}[t]
\centering
     \caption{\textbf{Generalization results on other backbone networks.} We evaluate the generalization performance of the FlowNetC and PWC-Net trained on different datasets: (A, C) FlyingChair, and (B, D) our dataset. (B, D) achieves better performance compared to (A, C). (E) is RAFT trained on our dataset as a reference.}
    \label{table5}
    \resizebox{\linewidth}{!}{
    \begin{tabular}{ccccccccc}
    \toprule
    \multirow{2}{*}{} & \multirow{2}{*}{Model} & \multirow{2}{*}{Dataset} & 
    \multicolumn{1}{c}{Sintel C.} & \multicolumn{1}{c}{Sintel F.} & \multicolumn{2}{c}{KITTI12} & \multicolumn{2}{c}{KITTI15} \\
    \cmidrule(){4-4}\cmidrule(lr){5-5}\cmidrule(lr){6-7}\cmidrule(lr){8-9}
    & & & EPE & EPE & EPE & F1 & EPE & F1 \\
    \midrule
    (A) & FlowNetC & Ch & 5.17 & 6.43 & 11.82 & 57.67 & 20.65 & 62.91 \\
    
    (B) & FlowNetC & Ours & \textbf{4.48} & \textbf{6.07} & \textbf{10.64} & \textbf{52.72} & \textbf{18.53} & \textbf{55.15} \\
    \midrule
    (C) & PWC-Net & Ch & 3.25 & 4.36 & 6.27 & \textbf{27.18} & 14.22 & 40.38 \\
    
    (D) & PWC-Net & Ours & \textbf{2.94} & \textbf{4.29} & \textbf{5.26} & 27.28 & \textbf{10.61} & \textbf{38.63} \\
    \midrule
    (E) & RAFT & Ours & 1.98 & 3.85 & 3.63 & 20.00 & 7.17 & 29.24\\
    \bottomrule
    \end{tabular}}
\end{table}

\paragraph{Competing Datasets for Training} We choose FlyingChairs (Ch)~\cite{flownet} and dCOCO~\cite{dcoco} as the competing datasets, and leverage the RAFT networks pre-trained on each dataset provided by the authors and dCOCO.
For multi-stage training models, from the networks pre-trained on ours, we further train with FlyingThings3D (Th)~\cite{things} in sequence to compare with the RAFT model trained with FlyingChairs followed by FlyingThings3D (Ch$\rightarrow$Th). 

\paragraph{Test Datasets} 
We evaluate on Sintel~\cite{sintel} and KITTI 2015~\cite{kitti15}.
These datasets contain crucial real-world effects, such as occlusions, illumination changes, motion blur, and camera noise, making them challenging and widely used standard benchmarks for evaluating optical flow models.
We report the performance of the model trained with the base datasets without fine-tuning on Sintel or KITTI, called \emph{generalization} and that of the model fine-tuned on the training set of Sintel or KITTI, called \emph{fine-tuning}. 

\paragraph{Evaluation} Following the convention,
we report the average End-Point Error (EPE) and the errors that exceed $3$ pixels and $5$\% of its true value (Fl). 
We further evaluate the percentage of pixels with an absolute error smaller or equal to $1$ ($\leq$1). 
The \textbf{bold} will be used to highlight the best one among the methods.

\subsection{Comparison with Other Synthetic Datasets} \label{subsection5} 
We compare the generalization and fine-tuning performance of the networks trained on our dataset and other competing datasets~\cite{flownet,things,dcoco}. 
For fair comparisons, we train the network on our dataset (denoted as Ours) with $20$k 
image pairs that include translation, rotation, and zooming. We also evaluate our dataset with occlusion masks $\langle$O$\rangle$ (denoted as Ours+O).

\paragraph{Generalization}
The left part of Table~\ref{table2} summarizes the generalization test.
Among the models trained on a single dataset, our datasets (C, D) show the best performance on Sintel.
However, dCOCO (B) shows better performance on KITTIs. 
We further evaluate the performance on two other benchmarks as shown in Table~\ref{table3}, and observe that dCOCO achieves better performance on Virtual KITTI~\cite{vkitti}, which is a synthetic dataset. On the other hand, ours achieves more accurate optical flow estimation in a real dataset, \ie, HD1K~\cite{HD1k}.
From these results, we assume that dCOCO, which uses depth-aware data generation approach with real images, is effective in autonomous driving scenarios and the similar motion distribution and texture between the synthetic and target dataset are key factors of generalization. 
We also pre-train the network on 2D motion datasets, such as FlyingChairs~\cite{flownet} and our datasets, and sequentially train on FlyingThings3D~\cite{things}. Compared to (E) which uses FlyingChairs at the initial stage, (F, G) show better generalization performance in the KITTIs and Sintel Clean pass. 
These show that the choice of the initial training stage significantly affects the final performance.

\begin{figure*}[t]
    \footnotesize
    \centering
    \resizebox{0.40\textwidth}{!}{
    \begin{tabular}
    {ccccccccc}
    \toprule
    \multirow{2}{*}{} &
    \multicolumn{2}{c}{Sintel C.} & \multicolumn{2}{c}{Sintel F.} & \multicolumn{2}{c}{KITTI12} & \multicolumn{2}{c}{KITTI15} \\ 
    \cmidrule(){2-3}\cmidrule(lr){4-5}\cmidrule(lr){6-7}\cmidrule(lr){8-9}
    & EPE & $\leq$1& EPE & $\leq$1 & EPE & Fl & EPE & Fl \\
    \midrule
    (A) & 3.39 & 0.72 & 5.68 & 0.66 & 8.56 & 44.05 & 14.11 & 48.20 \\ 
    (B) & \underline{2.63} & \underline{0.73} & \underline{4.33} & \underline{0.69} & \underline{6.95} & \underline{38.21} & \underline{12.57} & \underline{43.46} \\
    (C) & \textbf{2.55} &  \textbf{0.75} & \textbf{4.16} & \textbf{0.71} & \textbf{5.74} & \textbf{35.0} & \textbf{10.31} & \textbf{41.86}  \\
    \bottomrule
    \end{tabular}
    }
    \resizebox{0.594\textwidth}{!}{
    \begin{tabular}{@{}c@{}@{}c@{}@{}c@{}}
    \includegraphics[width=0.3\linewidth]{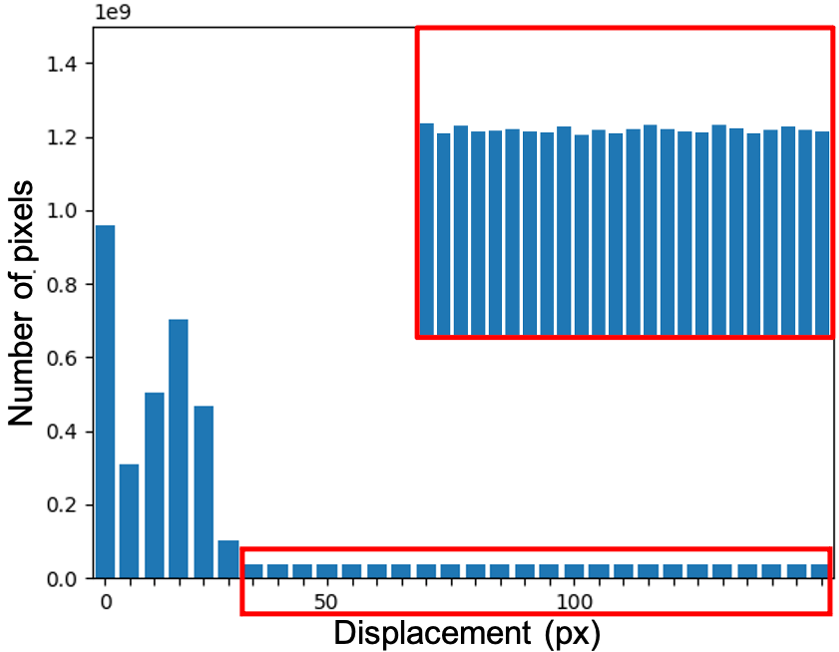}&
    \includegraphics[width=0.3\linewidth]{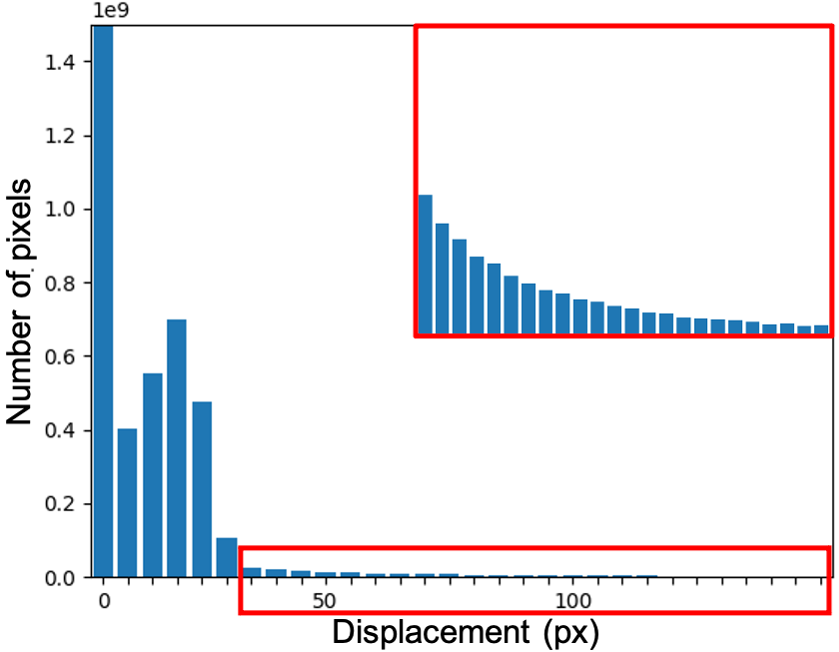}&
    \includegraphics[width=0.3\linewidth]{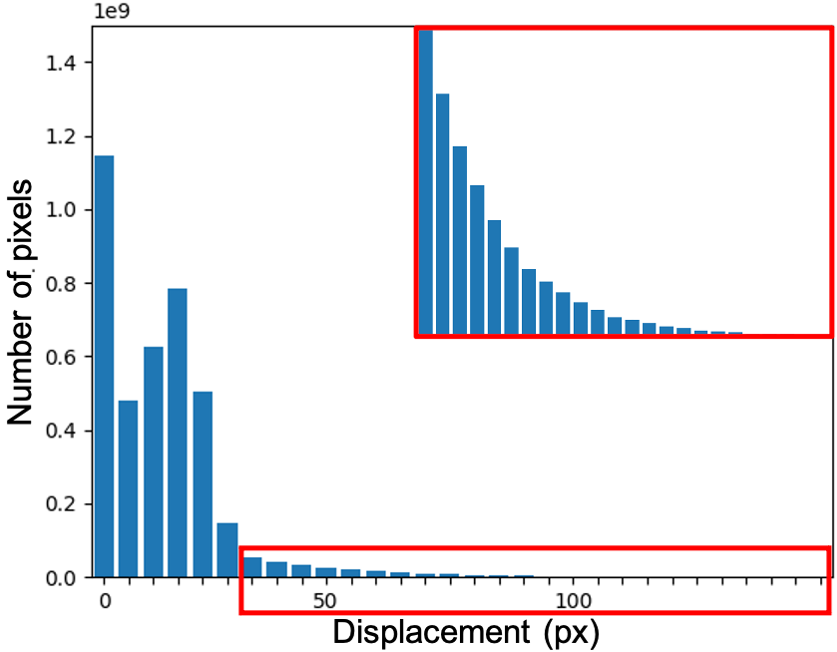}\\
    (A)&
    (B)&
    (C)
    \end{tabular}
    }
    \caption{
    \textbf{
    Generalization results and histograms of datasets depending on foreground translation distribution.} 
    From left to right, (A) uniform, (B) Gaussian, and (C) exponential distribution.
    (A) is sampled from a uniform distribution of the interval [0,150].
    (B) is the suggested distribution by FlyingChairs~\cite{flownet} given as $\mathrm{max}(\mathrm{min}(\mathrm{sign}(\gamma)\cdot|\gamma|^3,150),-150)$, where $\gamma \sim \mathcal{N}(0,\,2.3^2)$. 
    (C) is the proposed distribution that follows natural statistics~\cite{butler2012naturalistic}.
    Note that we sample foreground translation magnitude from the three distributions while the background distribution is fixed.
    }\label{figure:3}
    \vspace{-3mm}
\end{figure*}

\paragraph{Fine-tuning} 
We fine-tune the networks of the left part of Table~\ref{table2} on Sintel or KITTIs, and the results are reported in the right part of the table. Overall, our datasets show favorable performance.
Compared to (E) first pre-trained on FlyingChairs, (F, G) show better performance.
(G) especially achieves the lowest Fl and noticeable performance improvement in KITTI 2015. 
These results suggest that utilizing occlusion masks as a gradient suppression tool is effective in fine-tuning real-world datasets, \ie, KITTI 2012 and KITTI 2015. We observe a consistent tendency with the online benchmark results as follows.

\begin{table*}[t]
\centering
    \caption{\textbf{Impact of motion complexity and occlusion masks.} We evaluate the generalization and fine-tuning performance of the RAFT networks trained on our datasets with different motion types and occlusion masks: (A) translation, (B) adding rotation, (C) adding zooming, and (D) applying occlusion masks. We provide the performance of the network trained on Flyingchairs~\cite{flownet} (E) for the comparison.}
    \label{table6}
    \resizebox{0.85\linewidth}{!}{
    \begin{tabular}{@{}c@{}@{}c@{}@{}c@{}}
    \begin{tabular}{cc@{\hspace{0.5cm}}}
    \toprule
    &\multicolumn{1}{c}{}\\[5pt]
    &\multicolumn{1}{c}{}\\[5pt]
    & Motion types \\
    \hline
    (A) & T \\
    (B) & T+R \\
    (C) & T+R+Z \\
    (D) & T+R+Z+O \\
    \hline
    (E) & Ch \\
    \bottomrule
    \end{tabular}
    &
    \begin{tabular} {c@{\quad}cc@{\quad}cc@{\quad}cc@{\quad}c@{\hspace{0.5cm}}}
    \toprule
    \multicolumn{8}{c}{Generalization test}\\
    \cmidrule(lr){1-8}
    \multicolumn{2}{c}{Sintel C.} & \multicolumn{2}{c}{Sintel F.} & \multicolumn{2}{c}{KITTI12} & \multicolumn{2}{c}{KITTI15} \\
    \cmidrule(lr){1-2}\cmidrule(lr){3-4}\cmidrule(lr){5-6}\cmidrule(lr){7-8}
    EPE & $\leq$1 & EPE & $\leq$1 & EPE & Fl & EPE & Fl\\
    \hline
    2.55 &  0.75 & 4.16 & 0.71 & 5.74 & 35.69 & 10.31 & 41.86 \\
    2.33 & 0.81 & 4.01 & 0.77 &  6.09 & 30.71 & 12.09  & 38.82 \\
    \textbf{1.98} &  \textbf{0.86} & 3.85 & \textbf{0.82} & \textbf{3.63} & 20.00 & \textbf{7.17} & 29.24 \\
    2.02 & \textbf{0.86} & \textbf{3.67} & \textbf{0.82} & 3.66 & \textbf{19.37} & 7.88 & \textbf{28.41} \\
    \hline
    2.28 & 0.79 & 4.51 & 0.72 & 4.66 & 30.54 & 9.85 & 37.56 \\
    \bottomrule
    \end{tabular}
    &
    \begin{tabular}{c@{\quad}cc@{\quad}cc@{\quad}cc@{\quad}c}
    \toprule
    \multicolumn{8}{c}{Fine-tuning test}\\
    \cmidrule(lr){1-8}
    \multicolumn{2}{c}{Sintel C.} & \multicolumn{2}{c}{Sintel F.} & \multicolumn{2}{c}{KITTI12} & \multicolumn{2}{c}{KITTI15} \\
    \cmidrule(lr){1-2}\cmidrule(lr){3-4}\cmidrule(lr){5-6}\cmidrule(lr){7-8}
    EPE & $\leq$1 & EPE & $\leq$1 & EPE & Fl & EPE & Fl \\
    \hline
    0.93 & 0.93 & 1.45 & 0.89 & 1.46 & 5.09 & 2.49 & 8.71 \\
    0.90 & 0.93 & 1.40 & 0.89 & 1.44 & 4.82 & 2.53  & 8.84 \\
    \textbf{0.85} &  \textbf{0.94} & 1.40 &  \textbf{0.89} & \textbf{1.33} & 4.37 & 2.20 & 8.19 \\
    0.89 & 0.93 &  \textbf{1.39} &  \textbf{0.89} & 1.35 & \textbf{4.36} & \textbf{2.15} & \textbf{7.60} \\ 
    \hline
    0.89 & 0.93 & 1.49 &  \textbf{0.89} & 1.39 & 4.69 & 2.36 & 8.43 \\ 
    \bottomrule
    \end{tabular}
    \end{tabular}
    }
\end{table*}

\paragraph{Online Benchmarks}
We follow the training procedure described in RAFT~\cite{raft} to fine-tune the model pre-trained by our dataset and test on the public benchmarks of Sintel and KITTI15. 
As summarized in Table~\ref{table4}, using our dataset for the initial curriculum outperforms the original RAFT on both public benchmarks. On the KITTI15 test set, the network pre-trained on our synthetic dataset with occlusion masks shows better performance compared to RAFT. 
In the Sintel test dataset, we observe that the performance improvement in Sintel Clean and Final passes with our dataset. With and without the \textit{warm-start} initialization, the network trained with our training schedule also achieves better results in both passes. From these results, we assume that learning the simplest characteristics for estimating optical flow at the initial learning schedule without occlusion estimation helps the network perform better.

\paragraph{Other Backbone Networks}
To evaluate the effectiveness of our dataset other than RAFT, we selected two more optical flow models: FlowNet~\cite{flownet} and PWC-Net~\cite{pwc}.
We use the re-implementation of FlowNet~\footnote{https://github.com/ClementPinard/FlowNetPytorch} and PWC-Net~\footnote{https://github.com/visinf/irr}. 
Table~\ref{table5} shows the result of each network trained on our dataset outperforming the one trained on FlyingChairs~\cite{flownet}. We also contain the previous experiment with RAFT in (C) as a reference. These results prove that the simple properties of our dataset are effective for not only the RAFT~\cite{raft}, but also general optical flow networks.

\subsection{Ablation Study}\label{subsection4}
By virtue of the fast generation speed from the simple recipes and the controllability of our dataset, we can conduct a series of ablation studies to determine the critical factors of our dataset which affect the network performance the most.

\begin{table}[t]
\centering
     \caption{\noindent \textbf{Impact of abundant texture.} We train RAFT with a different number of foregrounds with/without applying gaussian blur. We provide the performance of the network trained on FlyingChairs~\cite{flownet} (D) for the comparison.}
    \label{table:7}
    \resizebox{1\linewidth}{!}{
    \begin{tabular}{cccccccccccc}
    \toprule
    &
    &
    Number of
    & 
    &
    \multicolumn{2}{c}{Sintel C.} & \multicolumn{2}{c}{Sintel F.} & \multicolumn{2}{c}{KITTI12} & \multicolumn{2}{c}{KITTI15} \\
    \cmidrule(lr){5-6}\cmidrule(lr){7-8}\cmidrule(lr){9-10}\cmidrule(lr){11-12}
    & Dataset & foregrounds & Blur & EPE & $\leq$1 & EPE & $\leq$1 & EPE & Fl & EPE & Fl \\
    \midrule
    (A) & Ours& 4 & \xmark &2.29 &  0.85 & 3.88 &  \textbf{0.82} & 3.95 & 21.04 & 8.84 & 30.96\\
    (B) & Ours& 8 & \xmark & \textbf{2.17} & \textbf{0.86} & \textbf{3.69} & \textbf{0.82} & \textbf{3.57} & \textbf{18.86} & \textbf{7.34} &  \textbf{28.09} \\
    (C) & Ours& 8 & \cmark &2.25 &  \textbf{0.86} & 3.70 & \textbf{0.82} & 3.91 & 21.87 & 8.96 &  31.05 \\
    \midrule
    (D) & Ch  & - & - & 2.28 & 0.79 & 4.51 & 0.72 & 4.66 & 30.54 & 9.85 & 37.5 \\
    \bottomrule
    \end{tabular}
    }
\end{table}

\paragraph{Foreground Translation Distributions} \label{dist} 
We evaluate the effect of the translational motion distribution of foregrounds with 20K image pairs.
We use three different distributions to sample magnitudes of translation.
Figure~\ref{figure:3} shows the histograms of each dataset distribution and summarizes the generalization results achieved by the RAFT network. (A) is uniform distribution and (B) is Gaussian distribution suggested by FlowNet~\cite{flownet}. (C) is the proposed distribution that follows natural statistics~\cite{butler2012naturalistic}.

As shown in the histograms, peaks are
near zero (in a factor of $10^9$) due to the background translation. Thus, we focus on the tails of the distributions, which typically occur by foregrounds. (A) includes excessively large motions, which are unrealistic in real-world scenarios and eventually degrade the performances.
Comparing with (B), (C) outperforms on overall metrics of benchmarks. 
The main difference between these two is the density of the focused region in the histogram, where (C) decays faster than (B).
From this, we observe that slight differences in tails of translation distributions affect the performance of the model significantly; thus, we take special care of a balanced motion distribution design.
We choose (C) as the distribution of translation for the following experiments.

\paragraph{Motion Complexity} We measure the effect of each motion type in training. Starting from the dataset having the translation $\langle$T$\rangle$ only, we sequentially apply rotation $\langle$R$\rangle$ and zooming $\langle$Z$\rangle$. As shown in (B) of Table~\ref{table6}, adding rotation transformation to (A) lowers the EPE on Sintel while increasing on KITTI. 
We observe that KITTI consists of driving scenes, where there are rare rotation motions. 
On the other hand, Sintel is a cinematic dataset including rotation motions caused by objects and cameras.
This implies that adding rotation might confuse the network on the test datasets that contain few rotation motions.
Interestingly, both (A) and (B) show comparable performance to the network trained on FlyingChairs (E), which contains three motion types, T+R+Z.
We regard that different translation distributions and abundant textures led to these results.
Finally, by adding zooming (C), the generalization performance outperforms (A), (B), and (E) in all cases.
We observe that zooming mimics the backward and forward object or camera motions, which frequently happens in both benchmarks. 
Hence, this may hint that the effect of adding rotation motion depends on the characteristics of test datasets, while zooming acts as more important factor for generalization.

The networks trained on our datasets have not seen any 3D motion during training; thus, we can further fine-tune on another dataset, including 3D motions in practice.
To figure out the ability of our datasets as pre-training datasets, we further fine-tune the aforementioned networks to the benchmarks, KITTI 2015 or Sintel.
We follow the same fine-tuning protocol suggested by Aleotti~\emph{et~al.}~\cite{dcoco} on the KITTI datasets.
The fine-tuning results in the right part of Table~\ref{table6} show a consistent tendency with the above generalization study.
While the improvement is marginal due to the high accuracy regime, the best performance is achieved when zooming is included in pre-training.
This suggests that zooming motion is challenging for the network to learn in fine-tuning before the pre-training in the curriculum learning sense~\cite{curriculum}.
We conclude that zooming is the most crucial factor among motion types of the synthetic dataset, improving the performance both in generalization and fine-tuning.

\paragraph{Effects of Occlusion Mask} The prior works~\cite{occlusion3, matters,superslomo} show the effectiveness of occlusion masks $\langle$O$\rangle$. 
Unlike these prior arts, we propose an intuitive and effective method utilizing the easily obtainable occlusion masks by suppressing the gradients at the regions to be occluded in a supervised manner.
In the left part of Table~\ref{table6}, generalization results with occlusion mask (D) show comparable EPE to (C) on the benchmarks but lower Fl on the KITTI datasets.
To further evaluate, we fine-tune the network (D) from the left part of Table~\ref{table6} on the benchmarks and show its results in the right part of the table.
The results also show lower Fl on the KITTI dataset. Besides, (D) outperforms (C) on both metrics in fine-tuning on KITTI 2015, which contains the most complicated real-world scenes. 
This shows that focusing on the areas that can be clearly learned from synthetic data helps networks learn complex effects, e.g., occlusion handling, real-world effects, and 3D motion in complicated scenes.
We observe the consistent tendency with the results of multi-stage training and public benchmarks as shown in Table~\ref{table4}.
This phenomenon can be regarded as curriculum learning, where learning more concepts gradually to complex one helps the network perform better. Applying the occlusion mask is an intuitive method for curriculum learning, and we proved the high effectiveness in improving the final performance.

\paragraph{Abundant Textures}
We analyze the effect of the abundant textures of foregrounds in training. Considering that the average number of foregrounds in the FlyingChairs~\cite{flownet} is 5, we compared the case when the number of foregrounds is 4 and 8. We also apply a Gaussian filter whose kernel size is 5 to the foregrounds for simulating the lack of high-frequency textures of chairs used in FlyingChairs. 
Table~\ref{table:7} shows that more foregrounds with high-frequency textures lead to overall  improvement. These results hint 
that abundant textures are another important factor
in generating synthetic data.

\section{Conclusion}
We propose an easily controllable synthetic dataset recipe by cut-and-paste, which enables conducting comprehensive studies. Through the experiments, 
we reveal the simple yet crucial factors for generating synthetic datasets and learning curriculums. We introduce a supervised occlusion mask method, which stops the gradient at the regions to be occluded.
Combining these findings, we observe that the networks trained on our datasets achieve favorable generalization performance, and our datasets with occlusion masks serve as a powerful initial curriculum, which achieves superior performance in fine-tuning and online benchmarks.

\bibliographystyle{ieeetr}
\bibliography{egbib}

\begin{thebibliography}{10}

\bibitem{flownet2}
E.~Ilg, N.~Mayer, T.~Saikia, M.~Keuper, A.~Dosovitskiy, and T.~Brox, ``Flownet
  2.0: Evolution of optical flow estimation with deep networks,'' in {\em IEEE
  Conference on Computer Vision and Pattern Recognition (CVPR)},
  pp.~2462--2470, 2017.

\bibitem{pwc}
D.~Sun, X.~Yang, M.-Y. Liu, and J.~Kautz, ``Pwc-net: Cnns for optical flow
  using pyramid, warping, and cost volume,'' in {\em IEEE Conference on
  Computer Vision and Pattern Recognition (CVPR)}, pp.~8934--8943, 2018.

\bibitem{sun2019models}
D.~Sun, X.~Yang, M.-Y. Liu, and J.~Kautz, ``Models matter, so does training: An
  empirical study of cnns for optical flow estimation,'' {\em IEEE Transactions
  on Pattern Analysis and Machine Intelligence (TPAMI)}, vol.~42, no.~6,
  pp.~1408--1423, 2019.

\bibitem{liteflownet}
T.-W. Hui, X.~Tang, and C.~C. Loy, ``Liteflownet: A lightweight convolutional
  neural network for optical flow estimation,'' in {\em IEEE Conference on
  Computer Vision and Pattern Recognition (CVPR)}, pp.~8981--8989, 2018.

\bibitem{raft}
Z.~Teed and J.~Deng, ``Raft: Recurrent all-pairs field transforms for optical
  flow,'' in {\em European Conference on Computer Vision (ECCV)}, Springer,
  2020.

\bibitem{flownet}
A.~Dosovitskiy, P.~Fischer, E.~Ilg, P.~Hausser, C.~Hazirbas, V.~Golkov, P.~Van
  Der~Smagt, D.~Cremers, and T.~Brox, ``Flownet: Learning optical flow with
  convolutional networks,'' in {\em IEEE International Conference on Computer
  Vision (ICCV)}, pp.~2758--2766, 2015.

\bibitem{dcoco}
F.~Aleotti, M.~Poggi, and S.~Mattoccia, ``Learning optical flow from still
  images,'' in {\em IEEE Conference on Computer Vision and Pattern Recognition
  (CVPR)}, 2021.

\bibitem{things}
N.~Mayer, E.~Ilg, P.~Hausser, P.~Fischer, D.~Cremers, A.~Dosovitskiy, and
  T.~Brox, ``A large dataset to train convolutional networks for disparity,
  optical flow, and scene flow estimation,'' in {\em IEEE Conference on
  Computer Vision and Pattern Recognition (CVPR)}, pp.~4040--4048, 2016.

\bibitem{sintel}
D.~J. Butler, J.~Wulff, G.~B. Stanley, and M.~J. Black, ``A naturalistic open
  source movie for optical flow evaluation,'' in {\em European Conference on
  Computer Vision (ECCV)}, pp.~611--625, Springer, 2012.

\bibitem{mayer2018makes}
N.~Mayer, E.~Ilg, P.~Fischer, C.~Hazirbas, D.~Cremers, A.~Dosovitskiy, and
  T.~Brox, ``What makes good synthetic training data for learning disparity and
  optical flow estimation?,'' {\em International Journal of Computer Vision
  (IJCV)}, vol.~126, no.~9, pp.~942--960, 2018.

\bibitem{kitti15}
M.~Menze and A.~Geiger, ``Object scene flow for autonomous vehicles,'' in {\em
  IEEE Conference on Computer Vision and Pattern Recognition (CVPR)},
  pp.~3061--3070, 2015.

\bibitem{horn_and_shcunk}
B.~K. Horn and B.~G. Schunck, ``Determining optical flow,'' {\em Artificial
  intelligence}, vol.~17, no.~1-3, pp.~185--203, 1981.

\bibitem{Black_M.J.}
M.~J. Black and P.~Anandan, ``A framework for the robust estimation of optical
  flow,'' in {\em 1993 (4th) International Conference on Computer Vision},
  pp.~231--236, IEEE, 1993.

\bibitem{TV-1}
C.~Zach, T.~Pock, and H.~Bischof, ``A duality based approach for realtime tv-l
  1 optical flow,'' in {\em Joint pattern recognition symposium}, Springer,
  2007.

\bibitem{discrete}
M.~Menze, C.~Heipke, and A.~Geiger, ``Discrete optimization for optical flow,''
  in {\em German Conference on Pattern Recognition}, pp.~16--28, Springer,
  2015.

\bibitem{Volumetric}
G.~Yang and D.~Ramanan, ``Volumetric correspondence networks for optical
  flow,'' {\em Advances in neural information processing systems}, vol.~32,
  2019.

\bibitem{costvolume}
J.~Xu, R.~Ranftl, and V.~Koltun, ``Accurate optical flow via direct cost volume
  processing,'' in {\em IEEE Conference on Computer Vision and Pattern
  Recognition (CVPR)}, pp.~1289--1297, 2017.

\bibitem{Iterative_residual}
J.~Hur and S.~Roth, ``Iterative residual refinement for joint optical flow and
  occlusion estimation,'' in {\em IEEE Conference on Computer Vision and
  Pattern Recognition (CVPR)}, pp.~5754--5763, 2019.

\bibitem{kitti12}
A.~Geiger, P.~Lenz, C.~Stiller, and R.~Urtasun, ``Vision meets robotics: The
  kitti dataset,'' {\em The International Journal of Robotics Research (IJRR)},
  vol.~32, no.~11, pp.~1231--1237, 2013.

\bibitem{slowflow}
J.~Janai, F.~Guney, J.~Wulff, M.~J. Black, and A.~Geiger, ``Slow flow:
  Exploiting high-speed cameras for accurate and diverse optical flow reference
  data,'' in {\em IEEE Conference on Computer Vision and Pattern Recognition
  (CVPR)}, pp.~3597--3607, 2017.

\bibitem{HD1k}
D.~Kondermann, R.~Nair, K.~Honauer, K.~Krispin, J.~Andrulis, A.~Brock,
  B.~Gussefeld, M.~Rahimimoghaddam, S.~Hofmann, C.~Brenner, {\em et~al.}, ``The
  hci benchmark suite: Stereo and flow ground truth with uncertainties for
  urban autonomous driving,'' in {\em Proceedings of the IEEE Conference on
  Computer Vision and Pattern Recognition Workshops}, pp.~19--28, 2016.

\bibitem{autoflow}
D.~Sun, D.~Vlasic, C.~Herrmann, V.~Jampani, M.~Krainin, H.~Chang, R.~Zabih,
  W.~T. Freeman, and C.~Liu, ``Autoflow: Learning a better training set for
  optical flow,'' in {\em IEEE Conference on Computer Vision and Pattern
  Recognition (CVPR)}, pp.~10093--10102, June 2021.

\bibitem{mm}
T.-H. Oh, R.~Jaroensri, C.~Kim, M.~Elgharib, F.~Durand, W.~T. Freeman, and
  W.~Matusik, ``Learning-based video motion magnification,'' in {\em European
  Conference on Computer Vision (ECCV)}, pp.~633--648, 2018.

\bibitem{voc}
M.~Everingham, L.~Van~Gool, C.~K. Williams, J.~Winn, and A.~Zisserman, ``The
  pascal visual object classes (voc) challenge,'' {\em International Journal of
  Computer Vision (IJCV)}, vol.~88, no.~2, pp.~303--338, 2010.

\bibitem{mscoco}
T.-Y. Lin, M.~Maire, S.~Belongie, J.~Hays, P.~Perona, D.~Ramanan,
  P.~Doll{\'a}r, and C.~L. Zitnick, ``Microsoft coco: Common objects in
  context,'' in {\em European Conference on Computer Vision (ECCV)},
  pp.~740--755, Springer, 2014.

\bibitem{roth2007spatial}
S.~Roth and M.~J. Black, ``On the spatial statistics of optical flow,'' {\em
  International Journal of Computer Vision (IJCV)}, vol.~74, no.~1, pp.~33--50,
  2007.

\bibitem{hur2019iterative}
J.~Hur and S.~Roth, ``Iterative residual refinement for joint optical flow and
  occlusion estimation,'' in {\em Proceedings of the IEEE/CVF Conference on
  Computer Vision and Pattern Recognition}, pp.~5754--5763, 2019.

\bibitem{jeong2022imposing}
J.~Jeong, J.~M. Lin, F.~Porikli, and N.~Kwak, ``Imposing consistency for
  optical flow estimation,'' in {\em Proceedings of the IEEE/CVF Conference on
  Computer Vision and Pattern Recognition}, pp.~3181--3191, 2022.

\bibitem{vkitti}
A.~Gaidon, Q.~Wang, Y.~Cabon, and E.~Vig, ``Virtual worlds as proxy for
  multi-object tracking analysis,'' in {\em IEEE Conference on Computer Vision
  and Pattern Recognition (CVPR)}, June 2016.

\bibitem{butler2012naturalistic}
D.~J. Butler, J.~Wulff, G.~B. Stanley, and M.~J. Black, ``A naturalistic open
  source movie for optical flow evaluation,'' in {\em European Conference on
  Computer Vision (ECCV)}, pp.~611--625, Springer, 2012.

\bibitem{curriculum}
Y.~Bengio, J.~Louradour, R.~Collobert, and J.~Weston, ``Curriculum learning,''
  in {\em International Conference on Machine Learning (ICML)}, pp.~41--48,
  2009.

\bibitem{occlusion3}
S.~Zhao, Y.~Sheng, Y.~Dong, E.~I. Chang, Y.~Xu, {\em et~al.}, ``Maskflownet:
  Asymmetric feature matching with learnable occlusion mask,'' in {\em IEEE
  Conference on Computer Vision and Pattern Recognition (CVPR)}, 2020.

\bibitem{matters}
R.~Jonschkowski, A.~Stone, J.~T. Barron, A.~Gordon, K.~Konolige, and
  A.~Angelova, ``What matters in unsupervised optical flow,'' in {\em European
  Conference on Computer Vision (ECCV)}, 2020.

\bibitem{superslomo}
H.~Jiang, D.~Sun, V.~Jampani, M.-H. Yang, E.~Learned-Miller, and J.~Kautz,
  ``Super slomo: High quality estimation of multiple intermediate frames for
  video interpolation,'' in {\em IEEE Conference on Computer Vision and Pattern
  Recognition (CVPR)}, pp.~9000--9008, 2018.

\end{thebibliography}

\end{document}